# Personalized News Recommendation System via LLM Embedding and Co-Occurrence Patterns

ZhengLi, Beijing University of Technology, lizhengcn@bjut.edu.cn

KaiZhang, State Key Laboratory of Communication Content Cognition, People's Daily Online,

zhangkai@people.cn

## ABSTRACT

In the past two years, large language models (LLMs) have achieved rapid development and demonstrated remarkable emerging capabilities. Concurrently, with powerful semantic understanding and reasoning capabilities, LLMs have significantly empowered the rapid advancement of the recommendation system field. Specifically, in news recommendation (NR), systems must comprehend and process a vast amount of clicked news text to infer the probability of candidate news clicks. This requirement exceeds the capabilities of traditional NR models but aligns well with the strengths of LLMs. In this paper, we propose a novel NR algorithm to reshape the news model via LLM Embedding and Co-Occurrence Pattern (**LECOP**). On one hand, we fintuned LLM by contrastive learning using large-scale datasets to encode news, which can fully explore the semantic information of news to thoroughly identify user preferences. On the other hand, we explored multiple co-occurrence patterns to mine collaborative information. Those patterns include news ID co-occurrence, Item-Item keywords co-occurrence and Intra-Item keywords co-occurrence. The keywords mentioned above are all generated by LLM. As far as we know, this is the first time that constructing such detailed Co-Occurrence Patterns via LLM to capture collaboration. Extensive experiments demonstrate the superior performance of our proposed novel method.

## 1 INTRODUCTION

Excellent recommendation algorithms play a huge role in solving information overload and improving user experience. In news platforms, NR algorithms provide users with personalized news. Those NR algorithms are ones that mines users' potential interests and preferences based on his/her historical click sequences and predicts the click probability of candidate news sets.

However, news is timely, which means news reports highlight recent events in order to quickly convey information to the public. That is why there are serious cold start problems in NR systems. Most of the news items that NR systems will face in the future will be ones that have never been seen before. Therefore, In NR field, our use of ID is not as flexible as in other sequential recommendation fields (such as movie recommendations, product recommendations). At the same time, in order to achieve accurate recommendations, news semantic mining is important. How to process news collaborative information and deeply mine news semantic information are two key issues for accurate prediction.

The Glove method, as a text encoding method, is widely used in NR. And it utilizes co-occurrence statistical information of words to learn vector representations of words. Inspired by that, we use LLM embedding combining co-occurrence patterns.

To solve the first problem mentioned above, we introduce these three co-occurrence patterns: news ID co-occurrence, Item-Item keywords co-occurrence and Intra-Item keywords co-occurrence. Specifically, if two items are adjacent in the item click sequence, it means that the keywords of the two items have a co-occurrence relationship, which means news ID co-occurrence relationship can be built. Then we use LLM to extract text keywords for each news. We use those keywords to represent an news item, and the keywords between and within adjacent items can be used respectively to build Item-Item keywords co-occurrence relationship and Intra-Item keywords co-occurrence relationship. The above relationships contains rich collaborative information, Several graphs can be constructed via those relationships and item embedding can be obtain by graph embedding algorithms. The collaborative information of each item can be obtained by integrating the embeddings of all co-occurrence patterns which represent this item. In order to solve the second problem mentioned above, we finetune LLM use a large scale of datasets including news datasets and several public datasets. Then we use LLM2vec[1] technology to encode the text of the item, so as to achieve the purpose of LLM to fully explore the semantics of the item text. Our main contributions are as follows:

- We first proposed to construct such detailed Co-Occurrence Patterns via LLM to capture collaboration. Through the relationship between adjacent items and keywords, we can construct news ID co-occurrence, Item-Item keywords co-occurrence and Intra-Item keywords co-occurrence. Three types of patent embeddings containing rich news items collaboration information.
- We propose to use the news corpus and public datasets to fine-tune the LLM via contrastive learning and use the LLM to obtain each news' embedding, so as to achieve the purpose of deeply mining the semantic information of news text.
- We conducted experiments on classic public news datasets (MIND[5]). Experimental results confirm that our method has significant improvements over state-of-the-art baselines.

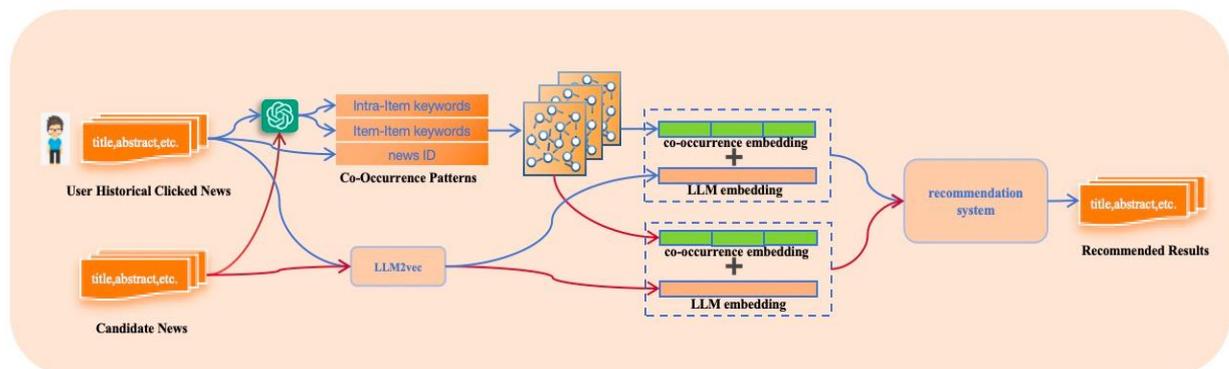

Fig.1 our proposed news recommendation via LLM Embedding and Co-Occurrence Pattern paradigm **LECOP**

## 2 RELATED WORK

### 2.1 NR

There are some classic NR algorithms. NPA[7] Uses the attention mechanism to establish item-level representation and user-level representation respectively. And CNN is used to learn the local context information of words. NAML[8] extends news level model to attentive multi-view learning (Title, Category, Content). LSTUR[9] upgrades the user-level model and introduces long-term and short-term user interest preference representation. NRMS[10] uses multi-head self-attention to capture long distance contexts of words instead of CNN. PPSR[11] uses hypergraphs to represent the complex relationships between titles, categories, and subcategories. At the same time, this work also uses popularity prediction to distinguish between obscure and popular news. When building item-level models, the above methods all use word-level attention mechanisms to obtain the final item representation. Glove[6] is generally used for word level representation due to its high efficiency. However, Glove is a static word embedding method, which means it cannot generate dynamic representations based on context.

2.2 LLM for NR

LKPNR[12] uses LLM embedding to mine news text semantic information and uses KG information to represent the collaborative relationship between items. However, LLM is not fine-tuned with news-related corpus and the KG entities are sparse so it performs poorly. Most importantly, it does not get rid of the traditional embedding method, but is only an additional. GNR leverages LLM to generate news theme and user interest. Then this method adds theme level representation to the news model, as well as add interest level representation to the user model. But as far as we know, it has not escaped the traditional embedding method either.

## 3 BACKGROUND

Background is presented in this section. We define $N_{click} = \{n_1, n_2, n_3, \cdots, n_m\}$ as the clicked news sequence, where $m$ is the number of click news items. Each news item contains the following information: newsid、title、abstract、category、subcategory and can be denoted as $n_i = (newsid_i, t_i, a_i, c_i, sc_i)$, where $newsid_i$, $t_i$, $a_i$, $c_i$, $sc_i$ representing separately newsid、title、abstract、category and subcategory. We define $N_{candidate} = \{n_1, n_2, n_3, \cdots, n_l\}$ as the candidate news, where $l$ is the number of candidate news items. Our task target is to predict the click propability of candidate news items.

## 4 PROPOSED MODEL

4.1 LLM2vec

We use LLM2vec[1] to get news embedding. We choose LLM2vec with supervised contrastive learning.

First, we fintune LLM2vec model using public datasets (E5 dataset) and the news dataset (MIND). To get training materials related with news, we reconstruct the MIND news dataset to the new format, One example is shown in Fig.2

> **Prompt:**
> "Given a news title, retrieve semantically similar abstract:"
> **Fintune data demo:**
> {"query": "5 Classic Appetizers That Make Holiday Hosting a Breeze", "positive": "Planning a celebration? These trusted "party starters" will please every single guest. Oh, and they couldn't be easier to make! The post 5 Classic Appetizers That Make Holiday Hosting a Breeze appeared first on Taste of Home.", "negative": "The Longhorns are 2-0 against K-State under Tom Herman."}

Fig.2 Finetune data sample for LLM2vec

As shown in Fig.2, we use retrieval logic to construct fine tuned data, using a news title as the "query", its abstract as the "positive" sample, and randomly selecting another news abstract as the "negative" sample. Then, we fintune LLM2vec model using above data materials.

Inspired by NoteLLM[2], when generating embeddings, we construct prompts in the following way: we constract prompt using all texts of each news item including title、abstract、category and subcategory. One example is shown in Fig.3

> "Given the information of a news, compress it into a maximum of 5 words for recommendation:"
> **category** : sports
> **subcategory** : football_nfl
> **title** : Should NFL be able to fine players for criticizing officiating?
> **abstract** : Several fines came down against NFL players for criticizing officiating this week. It's a very bad look for the league.

Fig.3 one prompt example of generating LLM embedding

To improve the effect, we used echo embedding[3], which can more deeply explore the semantic relationships in the context.

4.2 Keywords Co-occurrence

|  | News_ids in Test dataset | |
| --- | --- | --- |
|  | Duplicate Removal | Not Duplicate Removal |
| Not occur in Train dataset | 32.9% | 11.3% |
| Occur in Train dataset | 67.1% | 88.7% |

(a) News_ids dimension analysis

|  | Keywords in Test dataset | |
| --- | --- | --- |
|  | Duplicate Removal | Not Duplicate Removal |
| Not occur in Train dataset | 8.0% | 3.1% |
| Occur in Train dataset | 92.0% | 96.9% |

(b) Keywords dimension analysis

Table.1 The repeated occurrence of training set indicators in the test set

From Table.1, we can get the following conclusion: Regarding the repeated occurrence of training set indicators in the test set, the keyword dimension has a larger proportion of appearance compared to the news ID dimension. Therefore, studying the collaboration of keywords between news items is very meaningful.

Regarding the selection of LLM, we have chosen GLM4-9B-chat[4] to generate keywords by news titles and abstracts. Below is a prompt that uses news information to generate keywords:

**Prompt:**
"Given the information of a news, extract one to three keywords (The keywords must appear in the news information and must be nouns. Please provide the results in the following format: [keyword1, keyword2, keyword3]):"
+ news title + news abstract

Fig.4 a prompt to generate keywords for news

After generating keywords, in order to explore collaboration, we construct co-occurrence relationships using the historical click sequences of news items and keywords generated by LLM.

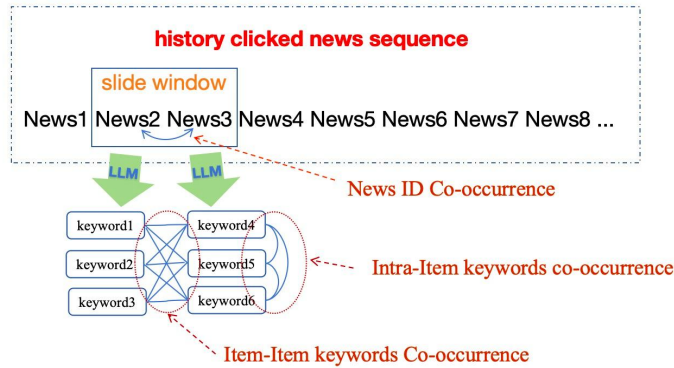

Fig.5 sliding window in the news click sequence to form co-occurrence relationships

We create a sliding window in the historical click sequence as shown in Fig.5. We believe that news items appearing simultaneously within this sliding window form co-occurrence relationships. In Fig.5, the sliding window size is set to 2. Besides, news ID co-occurrence, Item-Item keywords co-occurrence and Intra-Item keywords co-occurrence are obtained respectively. For example, "New2" and "New3" form news ID co-occurrence. "keyword1" and "keyword4" form Item-Item keywords co-occurrence, while "keyword4" and "keyword5" form Intra-Item keywords co-occurrence.

By utilizing the co-occurrence relationship between items or keywords, three weighted homogeneous graphs can be constructed. In each homogeneous graph, nodes are news items or keywords. Edges represent co-occurrence relationships, and the weights of edges represent co-occurrence frequency.

After that, we use graph representation learning method such as Node2Vec[13] to obtain the embedding of each nodes. We obtain node representations for news items and keywords respectively, and we refer to them as news ID co-occurrence embedding, Item-Item keywords co-occurrence embedding and Intra-Item keywords co-occurrence embedding. We get final co-occurrence embedding by concatenating the above 3 types of embeddings.

Finally, we combine final co-occurrence embedding with LLM embedding by adding directly to obtain the final news embedding. At this point, we complete the construction of the news model.

## 5 EXPERIMENTS

5.1 Dataset and Experimental Settings

In our experiment, we use MIND[5] dataset, which is shown in table 2. The MIND dataset is divided into two sizes, where the smaller one is obtained through random sampling from the larger one. Due to limited resources, in this paper, we mainly conduct experiments on the smaller one. The number of heads and the output dimension of each head are set to be consistent.

| datasets | User num | News num | Click num |
| --- | --- | --- | --- |
| MIND-small | 94057 | 65238 | 347727 |
| MIND-large | 1000000 | 161013 | 24155470 |

Table2　Comparative analysis of MIND-small and MIND-large

We trained with a batch size of 512 on a single 32GB Tesla V100 GPU. We chose a learning rate of 2e-4. We reduce the dimensionality of LLM embedding from 4096 to 300 through fully connected layers, so as to unify the dimensions of text embedding between the old and new methods. Adam optimizer is used to train the model. AUC, MRR, nDCG@5 and nDCG@10 are used as indicators for performance evaluation.

5.2 Performance Evaluation

| Methods | AUC | MRR | nDCG@5 | nDCG@10 |
| --- | --- | --- | --- | --- |
| NRMS | 0.6331 | 0.2923 | 0.3179 | 0.3825 |
| NRMS+LE | 0.6498 | 0.2993 | 0.3327 | 0.3969 |
| NRMS+LECOP | **0.6528** | **0.3071** | **0.3402** | **0.4036** |
| PPSR | 0.6932 | 0.3425 | 0.3813 | 0.4411 |
| PPSR+LE | 0.6912 | 0.3401 | 0.3758 | 0.4376 |
| PPSR+LECOP | **0.6941** | **0.3402** | **0.3770** | **0.4391** |

Table3 Experimental result

From the table3, it can be seen that our method(LECOP) achieved the best performance. From the comparative experiment of PPSR, it can be seen that when only LE (LLM embedding) is used, the optimal effect is not achieved, but with the addition of COP (Co-Occurrence Patterns), the effect reaches its optimum. Therefore, co-occurrence patterns are necessary.

## 6 CONCLUSION

In this paper, we introduce LLM embedding and Co-occurrence patterns into the NR system. First of all, to enhance semantic comprehension ability, we finetune LLM via contrast learning on a large scale datasets including MIND datasets and several public datasets. Then, to provide richer collaborative information, we propose three types of occurrence patterns including news ID co-occurrence, Item-Item keywords co-occurrence and Intra-Item keywords co-occurrence. Correspondingly, we can obtain three types of graphs separately. Using graph embedding, we obtain co-occurrence embedding. Finally, we combine LLM embedding and co-occurrence embedding in news model to replace traditional Glove semantic encoding in NR. Extensive experiments demonstrate the superiority of our novel method.

ACKNOWLEDGEMENT